\documentclass[letterpaper, 10 pt, conference]{ieeeconf}  

\IEEEoverridecommandlockouts                              

\usepackage{amsmath}
\usepackage{amssymb}
\usepackage{cite}
\usepackage{sdsmmMath}
\usepackage{mdpMath}
\usepackage{roboticsMath}
\usepackage{stemstyling}
\usepackage{graphicx}
\usepackage{subcaption}
\usepackage{pifont}
\usepackage{hyperref}
\usepackage{tabularx}
\usepackage{algorithm}
\usepackage{algpseudocode}
\usepackage[normalem]{ulem}

\DeclareMathOperator*{\argmin}{arg\,min}

\newcommand{\failureprob}{\pomdpGenProb{\text{failure}\vert x_{1:N_{points}}, a_{1:N_{actions}}, \pomdpBelief(\pomdpState_0)}}

\newcommand{\ALG}[0]{\textsc{RiskRL}}
\newcommand{\BASE}[0]{\textsc{BaseRL}}
\newcommand{\STT}[0]{$\texttt{st\_thomas}$}
\title{\LARGE \bf
When to Localize? \\A Risk-Constrained  Reinforcement Learning Approach
}

\author{Chak Lam Shek$^*$, Kasra Torshizi$^*$, Troi Williams, and Pratap Tokekar
\thanks{$^*$C. Shek and K. Torshizi contributed equally and are listed alphabetically.}
\thanks{This research was funded in part by the National Science Foundation (NSF) Eddie Bernice Johnson INCLUDES initiative, Re-Imagining STEM Equity Utilizing Postdoc Pathways (RISE UPP), award \#2217329. All authors are at the University of Maryland, College Park, MD 20742, USA.
        {\tt\small \{cshek1,ktorsh,troiw,tokekar\}@umd.edu}}
}

\begin{document}

\maketitle
\thispagestyle{empty}
\pagestyle{empty}


\begin{abstract}
In a standard navigation pipeline, a robot localizes at every time step to lower navigational errors. However, in some scenarios, a robot needs to selectively localize when it is expensive to obtain observations. For example, an underwater robot surfacing to localize too often hinders it from searching for critical items underwater, such as black boxes from crashed aircraft. On the other hand, if the robot never localizes, poor state estimates cause failure to find the items due to inadvertently leaving the search area or entering hazardous, restricted areas. Motivated by these scenarios, we investigate approaches to help a robot determine ``when to localize?'' We formulate this as a bi-criteria optimization problem: minimize the number of localization actions while ensuring the probability of failure (due to collision or not reaching a desired goal) remains bounded. In recent work, we showed how to formulate this active localization problem as a constrained Partially Observable Markov Decision Process (POMDP), which was solved using an online POMDP solver. However, this approach is too slow and requires full knowledge of the robot transition and observation models. In this paper, we present \ALG{}, a constrained Reinforcement Learning (RL) framework that overcomes these limitations. \ALG{} uses particle filtering and recurrent Soft Actor-Critic network to learn a policy that minimizes the number of localizations while ensuring the probability of failure constraint is met. Our numerical experiments show that \ALG{} learns a robust policy that leads to at least a 26\% increase in success rates when traversing unseen test environments.\\
Code: https://github.com/raaslab/when-to-localize-riskrl
\end{abstract}

\section{Introduction}\label{sec:introduction}

\begin{figure}[t]
     \centering

     \includegraphics[width=\linewidth]{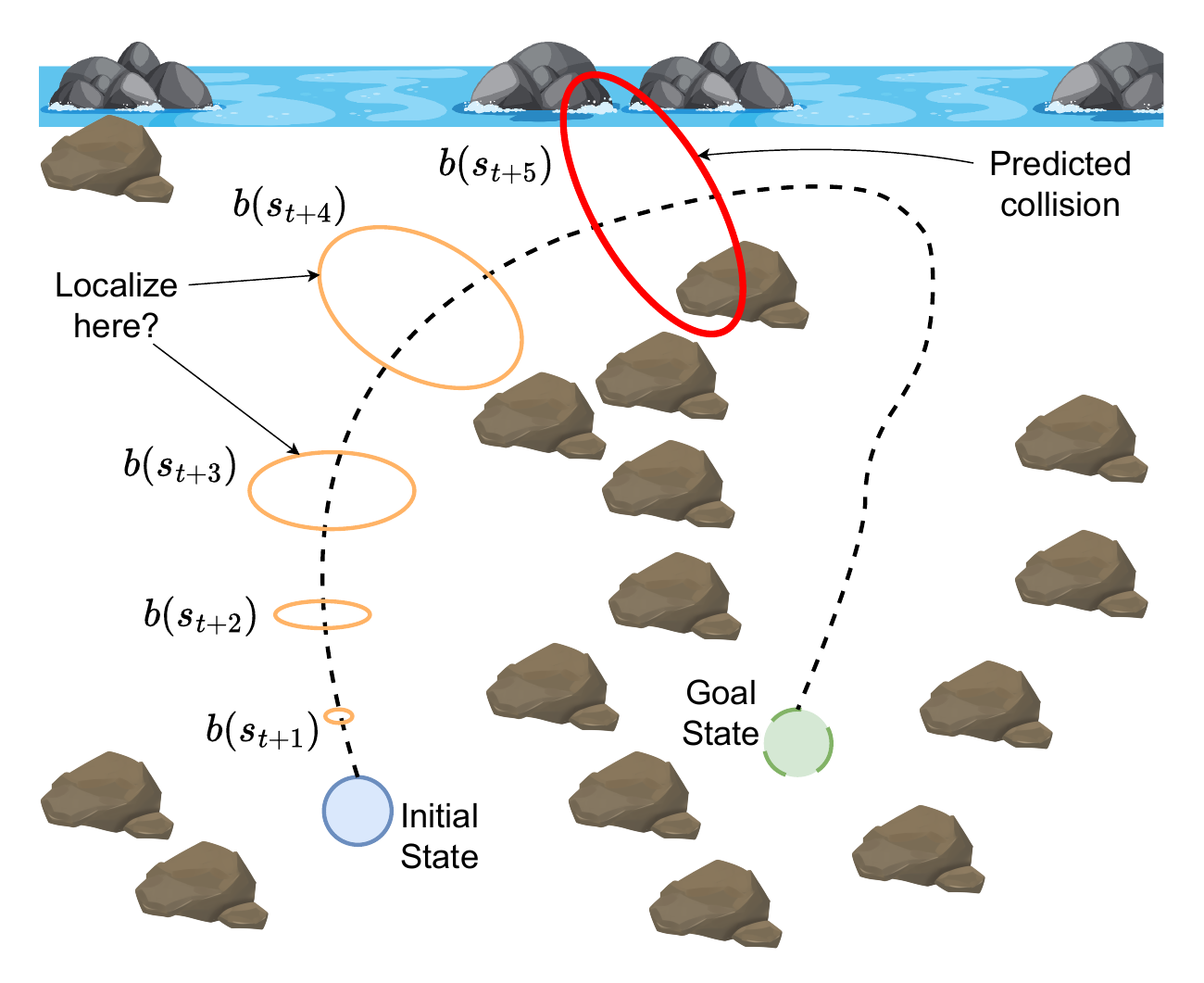}
     \caption{Motivating example. Consider a robot that may want to seldom localize (e.g., due to resource constraints) while traveling along a path (black dashed line). Despite obstacles (rocks and water), the autonomous robot can execute a series of open-loop motions for some period. However, as the dead reckoning uncertainty (gold and red ellipses) grows, the probability of failure (such as collision) may become too large. As such the robot must localize at some point to avoid failures. Thus, \textbf{our question} is: \textit{When should the robot localize to reduce failure probabilities?} }
     \label{fig:w2l_concept}
\end{figure}

In robotics, self-localization is crucial because it enhances navigation accuracy, and situational awareness and enables complex tasks. Typically, an autonomous robot perceives its environment and self-localizes, plans its subsequent actions, acts upon its plan phases, and repeats the cycle. However, sometimes, a robot may want to localize seldom when it is not advantageous. For example, underwater robots need to surface to localize in underwater rescue and recovery missions. Surfacing to localize too often may hinder an underwater robot from searching for critical underwater items such as black boxes from crashed aircraft. On the other hand, if the robot never localizes, it will accumulate large amounts of drift \cite{pereira2013risk}, which may prevent it from finding the items due to inadvertently leaving the search area or entering hazardous, restricted areas, as illustrated in Figure \ref{fig:w2l_concept}. Additionally, since the robot cannot execute movement and localization actions simultaneously, robots must balance prolonged actions that achieve mission objectives (such as searching for critical items underwater) with localizing to improve navigation accuracy.

We explored such scenarios in our recent work \cite{Williams2024w2l}. Our central question was: \emph{how can a robot plan and act for long horizons safely and only localize when necessary?} (Figure \ref{fig:w2l_concept} discusses a general scenario to this question). We emphasize that such a question is not trivial because we have two competing objectives. The first objective is \textit{localize often} to maximize mission safety and performance, where we can ensure the vehicle remains within the search area and out of hazardous zones.
On the other hand, the second objective is to \textit{localize infrequently} to minimize the number of times the vehicle must deviate from its mission, which in turn can reduce mission time. These two objectives are challenging to optimize via one objective, as shown with our POMCP baseline in \cite{Williams2024w2l}. Therefore, we addressed the question by formulating it as a constrained Partially Observable Markov Decision-making Process (POMDP), where our objective was to minimize the number of localization actions while not exceeding a given probability of failure due to collisions. Then we employed \texttt{CC-POMCP} \cite{Lee2018MonteCarloTS}, a cost-constrained POMDP solver, to find policies that determine when the robot should move along a given path \textit{or} localize.

Although our prior approach produced policies that reached the goal and outperformed baselines, the approach had limitations. First, \texttt{CC-POMCP} was computationally expensive, requiring over 20 minutes of inference to navigate a path of 55 waypoints. Second, \texttt{CC-POMCP} requires a well-defined model of the environment, including the robot's transition (motion) and observation models. Requiring such models may be problematic in unknown or dynamic real-world environments where it may be challenging to obtain accurate models. Finally, \texttt{CC-POMCP} is highly sensitive to transition noise, often failing to reach the goal as noise increases. As shown in Figure \ref{fig:num_and_success_train}, \ALG{} achieves a higher success rate under the same transition noise conditions, demonstrating greater robustness.

We propose a novel approach termed \ALG{} that employs constrained Reinforcement Learning (RL) \cite{9718160} and Particle Filters (PF) to overcome these limitations. Our new approach has multiple advantages over our prior work. First, although our new approach has a longer, single training time, it infers quicker during deployment, \textbf{enabling real-time planning}. The second was \textbf{reducing the need for accurate transition and observation models of the environment}, which we demonstrate by varying the transition and observation noises. In the formulation, the risk is modeled as a probability constraint, allowing us to design policies that minimize the failure probability while ensuring the robot remains within acceptable risk levels. This formulation provides greater control over the failure rate by explicitly incorporating risk constraints into the decision-making process. Furthermore, we use a PF to maintain the robot's belief as the robot executes noisy motion commands and receives noisy measurements from the environment. We also use the PF to compute the observation for the RL robot. 

We perform numerical experiments to compare \ALG{} with several baselines, including standard RL (\BASE{}), \texttt{CC-POMCP}, and heuristic policies. Our main finding is that when deployed in unseen testing environments, \ALG{} outperforms the \BASE{} and \texttt{CC-POMCP} baselines by at least \textbf{26\%} in terms of the success rate while also being the only algorithm that satisfies the risk constraint. Unlike \BASE{}, \texttt{CC-POMCP}, and the heuristics algorithm,  \ALG{} generalizes to our unseen test environments.

The remainder of this paper is organized as follows: Section II reviews the related work in active localization and constrained RL. Section III defines the problem formulation. Section IV presents the proposed \ALG{} framework. Section V provides experimental evaluations and comparisons with baselines. Finally, Section VI concludes the paper and discusses future directions.

\section{Related Work}

This paper explores minimizing localization actions while not exceeding pre-defined failure probabilities. In the following subsections, we position our method within the active localization and constrained RL literature.

\subsection{Active Localization}

Active perception \cite{Cowan1988AutomaticSensor,bajcsy2018revisiting} encompasses various approaches, including particle filter \cite{9267899, 10056178}, active localization \cite{Burgard1997ActiveLocalization, Borghi1998MinimumUncertainty}, active mapping \cite{Placed2023ActiveSLAMSurvey,Sasaki2020WhereToMap,Dhami2023PredNBV}, and active SLAM \cite{Placed2023ActiveSLAMSurvey}. These approaches focus on finding optimal robot trajectories and observations to achieve mission goals. Of these approaches, our current approach falls under active localization. Typically, active localization methods address \textit{where} a robot moves and looks to localize itself \cite{Mostegel2014ActiveMonocular, Otsu2018Where, Gottipati2019DAL, Strader2020PerceptionAware} or a target \cite{Tallamraju2020AirCapRL, Williams2023WhereAmI}. Thus, these active localization approaches generally differ from our problem because we seek to determine \textit{when} a robot localizes. However, one exception is our prior work \cite{Williams2024w2l}, which proposes an approach that precedes the one in this paper. This approach improves upon \cite{Williams2024w2l}, where we now model the probability of failure in terms of risk, reduce the inference time significantly, and relax the need for well-defined noise models.

\subsection{Constrained Reinforcement Learning}

Model-free reinforcement learning promises a more scalable and general approach to solving the active localization problem since it requires less domain knowledge~\cite{schulman2017proximalpolicyoptimizationalgorithms, mnih2016asynchronousmethodsdeepreinforcement}. However, many prior works applying RL to solving POMDPs, even without constraints, have gotten poorer results compared to more specialized methods \cite{9636140}. A recent architecture \cite{ni2022recurrentmodelfreerlstrong} utilizing a Soft-Actor Critic (SAC) with two separate recurrent networks for both the actor and value functions has shown promise to surpass more specialized methods in select examples. Since recurrent networks also act on the history of observations, they can handle partial observability. Our \ALG{} approach is based on this twin recurrent network SAC architecture~\cite{ni2022recurrentmodelfreerlstrong}. However, unlike \cite{ni2022recurrentmodelfreerlstrong} we also seek to deal with risk constraints. 

There is a separate line of work on constrained RL in the fully observable setting~\cite{pmlr-v139-amani21a}. Constrained RL extends model-free RL by incorporating real-world limitations, such as safety, budget, or resource constraints, into the learning process \cite{pmlr-v139-amani21a}. The primary challenge in constrained RL is balancing the trade-off between maximizing cumulative rewards and satisfying these secondary constraints. A commonly used approach to tackle this problem is the primal-dual optimization method \cite{liang2018acceleratedprimaldualpolicyoptimization}, which iteratively adjusts both the policy and constraint parameters. Various techniques are used to simplify and solve these problems, such as transforming the constraints into convex functions \cite{yu2019convergentpolicyoptimizationsafe} or employing stochastic approximation methods to handle probabilistic constraints \cite{9718160}. However, these prior works on constrained RL have only focused on the fully observable setting. In this paper, we build on these two lines of work and present \ALG{}, which handles both partial observability as well as chance constraints.

\begin{figure}
    \centering
    \includegraphics[width=\linewidth]{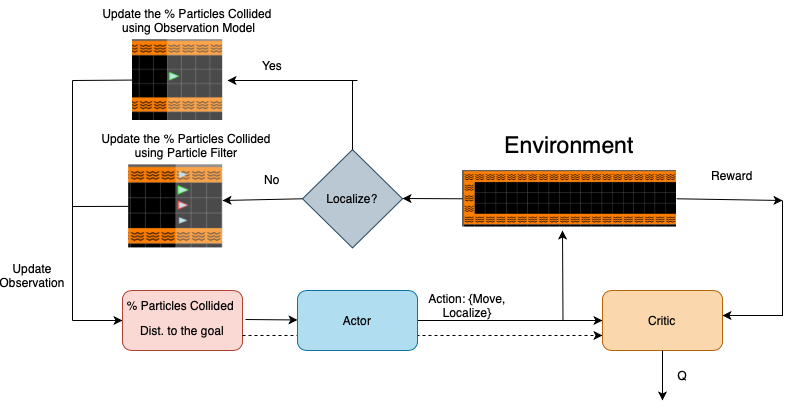}
    \caption{This flowchart depicts a decision-making process for a robot interacting with an environment. It combines a high-level (Soft Actor-Critic) RL planner for decision-making (Section \ref{sec:method:highlevel_planner}) and a Particle Filter for state estimation.
    }
    \label{fig:env_sys}
\end{figure}

\section{Problem Statement}\label{sec:formulation}

This paper solves the same active localization problem as in our prior work \cite{Williams2024w2l}. That is, a robot aims to selectively localize while navigating along a path to a pre-defined goal. When the robot believes it is opportune to localize, it uses its sensors to obtain noisy observations of its pose to mitigate failures such as collisions. Thus, we aim to generate a move-localize policy that 1) minimizes localization events and 2) avoids exceeding a failure probability threshold $\hat{c}$. For the reader's convenience, we include the original objective function from \cite{Williams2024w2l}:
\begin{equation}\label{eq:min_localize_policy}
\begin{split}
    \pomdpOptimalPolicy & = \argmin_{\pomdpPolicy\in\Pi} \sum_{t = 0}^{N_{actions}} \pomdpAction_t = \{\text{localize}\} \\
        \text{s.t.} & ~\failureprob \leq \hat{c},
\end{split}
\end{equation}
where $N_{actions}$ is the total number of actions, $a_{1:N_{actions}}$ is the sequence of move and localize actions, and $\pomdpBelief(\pomdpState_0)$ is the initial belief at the start of the path. Finally, $\Pi$ denotes the set of all possible action sequences over $N_{actions}$ timesteps.

\section{\ALG{} Active Localization Algorithm}\label{sec:method}

\begin{algorithm}
\caption{\ALG{} Active Localization Algorithm}
\label{alg:active_localize}
\begin{algorithmic}
\State Initialize RL planner state $s_t$, initial PF belief $b(s_0)$, path $x_{1:N_{points}}$
\While {not in a terminal state} 
    \State RL planner uses $s_t$ to select $a_t$ (move/localize)\;
    \If{$a_t = \text{move}$}
        \State Low-level planner computes motion command $u_t$
        \State Low-level planner truncates path $x_{2:N_{points}}$\;
        \State Robot executes $u_t$\;
        \State PF propagates belief $b(s_t)$ using $u_t$\;
    \EndIf
    \If {$a_t$ = localize}
        \State Robot receives observation $o_\text{state}$\; 
        \State PF updates belief $b(s_t)$ using $o_\text{state}$\;
        \State Low-level planner replans hazard-free path to goal\;
    \EndIf
    \State Robot receives reward $r_t$ \;
    \State PF computes $o_\text{planner}$ for the next time step\; 
\EndWhile
\end{algorithmic}
\label{algo:active_localize}
\end{algorithm}

We now present our algorithmic framework (Figure \ref{fig:env_sys}) for solving the active localization problem in \eqref{eq:min_localize_policy}. Our framework has three main components: a high-level planner, a low-level planner, and a PF. We discuss the high-level planner (which chooses when to move or localize) and our RL solution in Section \ref{sec:method:highlevel_planner}. The low-level planner selects a motion command $u_t$ if the agent wants to move, or localizes and replans its trajectory if it wants to localize. Finally, the PF provides observations $o_{planner}$ to the high-level planner and maintains the agent's belief using $u_t$ and noisy, 2D pose observations $o_{state}$ from the environment. Algorithm \ref{algo:active_localize} describes how our active localization algorithm works.

\subsection{High-Level Planning}\label{sec:method:highlevel_planner}

\textbf{Overview.} The high-level planner chooses \textit{which} high-level action the agent performs next:  $ a \in \mathbb{A}_\text{RLP} = \{\text{move}, \text{localize}\}$. The planner chooses the next action based on the 2D observation vector $o_{planner}$ from the PF. This vector contains the predicted collision probability $\hat{p}\in[0,1]$ and distance $\hat{d}\in[0, +\inf)$ to the goal. We compute $\hat{p}$ by counting the number of particles that collided with an obstacle and $\hat{d}$ as the number of steps between the belief's mean and the goal.

In this paper, we implement the high-level planner using heuristics for our baselines without learning and SAC neural networks for \ALG{} and \BASE{}. We describe the heuristics-based planners in Section \ref{sec:evaluations} and the \ALG{} planner below. To select the next high-level action $a_t$, the RL planners use the hidden state $h_t$ computed from the LSTM module, the previous action $a_{t-1}$, the observation vector $o_{planner}$,  while the heuristic planners either use $o_{planner}$ alone or internal data such as the number of moves between localize actions. To train the RL agent, the reward $r_t$ is chosen to be $ -1$ if the robot chooses the localize action and 0 otherwise. We chose this observation representation so that our RL planner generalizes to unseen environments.

\textbf{Chance-Constrained Planning (\ALG{}).} The original objective \eqref{eq:min_localize_policy} does not align with the standard RL formulation for expected discounted reward. Thus, we propose a relaxation, as shown in Equation \ref{eq:discounted_reward}. This relaxation allows us to use standard RL to minimize the number of localization actions by maximizing this reward function. However, naive optimization will violate the constraints in
(\ref{eq:min_localize_policy}).

The constraint probability is difficult to estimate because it requires interaction with the environment and varies based on the policy being used, making it challenging to establish a clear relationship between the policy and the constraint. We follow the relaxation approach outlined in \cite{9718160}, converting the probabilistic constraint in a Chance-Constrained POMDP into a cumulative constraint. Specifically, we reformulate the optimization problem as follows. Our goal is to maximize the expected cumulative reward \( V(\theta) \), defined by:
\begin{equation}
    \max_{\theta \in \mathbb{R}^d} V(\theta) \triangleq \mathbb{E} \left[ \sum_{t=0}^{\infty} \gamma^t r(s_t, a_t) \mid \pi_\theta \right]
    \label{eq:discounted_reward}
\end{equation}
where \( \theta \) denotes the parameters of the policy \( \pi_\theta \) and \( V(\theta) \) denotes the expected reward over time. We then impose a cumulative constraint:

\begin{equation}
    U_\theta : \sum_{t=0}^{T} \gamma^t (1 - \Pr(\text{failure} \mid x_{1:N}, a_{1:t}, b(s_0)))  \geq c,
\end{equation}
where \( U_\theta \) is the accumulated discounted probability of failure and \( c = \frac{(1 - \hat{c} \gamma^T(1 - \gamma))}{(1 - \gamma)} \) represents the risk-adjusted threshold. This formulation simplifies the original problem by approximating the probabilistic constraint, allowing us to evaluate the constraint based on the data generated from the rollout trajectory.

To incorporate the approximation of the probabilistic constraint into the reward function, we adjust the reward to account for the constraint violation. The new reward function is formulated as follows:
\begin{equation}
    \hat{r}(s_t, a_t) = r(s_t, a_t) + \lambda \left(I(s_t \notin \text{failure}) - c (1 - \gamma)\right), \label{eq:reward}
\end{equation}
where \( \hat{r}(s_t, a_t) \) represents the adjusted reward function that is $-1$ if the action is $\it{localize}$ and 0 otherwise, \( \lambda \) is a penalty coefficient, \( I(s_t \notin \text{failure}) \) is an indicator function that is 1 if the state \( s_t \) is not in the failure set and 0 otherwise, and \( c (1 - \gamma) \) is the threshold term derived from the relaxed constraint. The difference between the indicator function's value and the threshold can estimate the probability that the constraints are satisfied.

The algorithm \cite{9718160} shown below employs the primal-dual method to optimize the expected reward while satisfying constraints. The primal component focuses on maximizing the reward function as defined in Equation (\ref{eq:reward}), while the dual component adjusts the $\lambda$ values to control the risk levels associated with these constraints. It iteratively updates the policy parameters and dual variables by simulating trajectories and estimating gradients. The policy can be updated by computing the policy gradient defined by 
    ${\nabla_{\theta} L(\theta_k, \lambda_k) = \hat{r}(s_t, a_t) \nabla_{\theta} \log \pi_{\theta_k} (a_0 | s_0)}$,
where $\nabla_{\theta} \log \pi_{\theta_k}(a_0 | s_0)$ is the gradient of the log-probability of the policy $\pi_{\theta_k}$ selecting action $a_0$ given state $s_0$.

\begin{algorithm}[H]
\caption{Primal-Dual Optimization \cite{9718160}}
\label{alg:policy_optimization}
\begin{algorithmic}
\State Initialize $\theta_0$, $\lambda_0$, $T$, $\eta_\theta$, $\eta_\lambda$, $\delta$, $\epsilon$ 
\While {not converged} 
    \State Rollout a trajectory with the policy $\pi_{\theta_k}(s)$\; 
    \State Estimate primal gradient $\nabla_\theta L(\theta_k, \lambda_k)$ \; 
    \State Estimate dual gradient $U(\theta_k) - c$\; 
    \State Update primal variable: $\theta_{k+1} = \theta_k + \eta_\theta \nabla_\theta L(\theta_k, \lambda_k)$\; 
    \State Update dual variable: $\lambda_{k+1} = \lambda_k - \eta_\lambda (U(\theta_k) - c)$\; 
\EndWhile
\end{algorithmic}
\end{algorithm}

\begin{figure*}
    \centering
    \includegraphics[width=\linewidth]{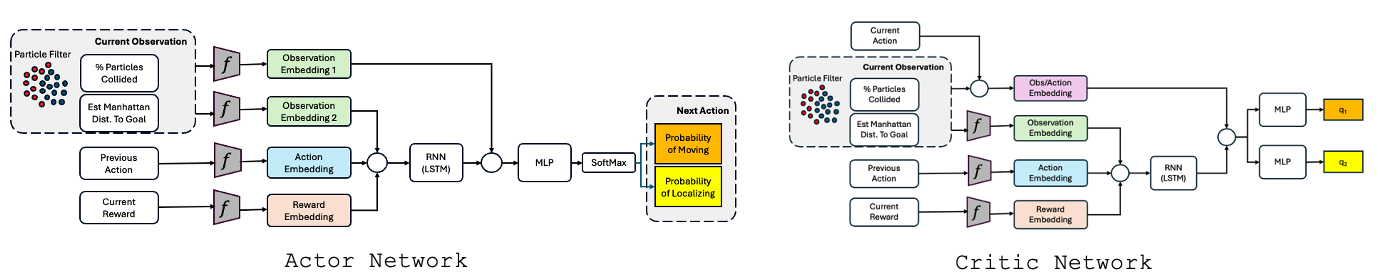}
    \caption{A diagram of the RL architecture used to train the models}
    \label{fig:rl_network}
\end{figure*}

\textbf{Handling Partial Observability.}
The previous section described how we can optimize the policy using the primal-dual approach. This section presents the specific architecture we use for the actual policy. 
Stemming from \cite{ni2022recurrentmodelfreerlstrong}, we use a Soft Actor-Critic Model as it generally tends to have better sample efficiency. To deal with partial observability, Recurrent Neural Networks (RNN) have been known to mitigate the effects of a noisy observation by making decisions based on the past trajectory instead of just the current observation (or fixed sequence of observations) \cite{knight2008stable}\cite{ho2017deep}. We implement an LSTM module to help stabilize training \cite{hochreiter1997long}. All of our embeddings are obtained with a one-layer MLP. Figure \ref{fig:rl_network} provides an illustrative diagram of the architecture.

\section{Experimental Evaluation}\label{sec:evaluations}
In this section, we report our findings from numerical experiments comparing \ALG{} with several baselines and evaluating the robustness and generalization capabilities of \ALG{}. The training and testing environment are motivated by an underwater scenario introduced in our prior work~\cite{Williams2024w2l}. In such work, a robot must balance localizing at the surface and navigating through underwater environments, which include obstacles such as rocks and coral formations, to search for items such as the black boxes of downed airplanes.

\subsection{Setup}
\begin{figure*}[]
    \centering
    \includegraphics[width=\linewidth]{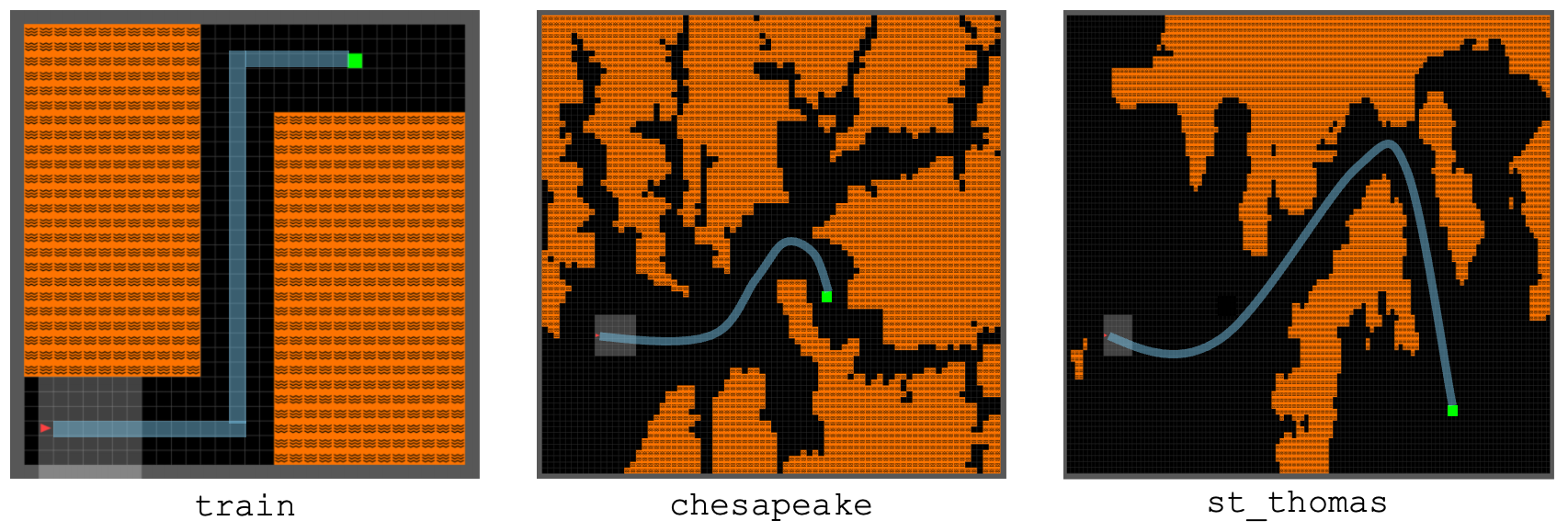}
    \caption{This figure shows our three evaluation environments in Minigrid. A red triangle denotes the agent, the orange squares denote the failure states (obstacles), and the sample paths are denoted in blue with the green squares representing the goal. We trained the RL agents using \texttt{train}.}
    \label{fig:environments}
\end{figure*}

\textbf{Baselines.} We compare \ALG{} with four types of baselines. The first type is two static policies (\texttt{SP}): \texttt{(M2x,L)}, and \texttt{(M3x,L)}, where \texttt{(M2x,L)} repeatedly moves twice and then localizes once. The second type is a threshold planner (\texttt{TP}) that localizes the robot whenever the collision probability exceeds a threshold. The third type is \texttt{CC-POMCP}, a cost-aware, online policy planner from our prior work \cite{Williams2024w2l}. Finally, the last type is \BASE{}, a standard, risk-unaware RL policy. For \BASE{}, we penalize the robot each time it localizes or collides (Table \ref{table:op_parameters}). This baseline is used to assess the advantage of employing risk-aware RL.

\begin{table*}

    \begin{tabularx}{1.003\textwidth} {|c c c c c c c c c c c c c|}
     \hline
      Planner & $\pomdpReward_{goal}$ & $\pomdpReward_{move}$ & $\pomdpReward_{local}$ & $\pomdpReward_{fail}$& \# Particles & $\ell_r$ & $\gamma$ & $\alpha$ & $\tau$ & DQN Layers & Policy Layers & Obs Emb. Size \\ [0.15ex] 
     \hline 
     \BASE & 0 & 0 & -1 & -256 & 100 & 0.00012 & 0.95 & 0.25 & 0.005 & [64, 64] & [64, 64] & 32 \\
     \ALG  & " & " & " & 0 & " & 0.0001 & 0.9 & 0.5 & " & [128, 128] & [128, 128] & 32 \\
     \hline
    \end{tabularx}
    \caption{RL Parameters. $\ell_r$ is the learning rate, $\gamma$ is the discount factor, $\alpha$ is the target entropy, and $\tau$ is the soft update.}
    \label{table:op_parameters}
\end{table*}

\textbf{Environments.} Figure \ref{fig:environments} shows our environmental setup. We assume the robot knows the map, the start state, and the goal region. We set the initial belief to the start state. In all of our experiments, we set the transition noise such that the robot has an 80\% chance of going forward and a 10\% chance of drifting to the left or the right. The observation noise is drawn from a \texttt{3x3} matrix with a 68\% probability of observing its true position and a 4\% probability of observing a neighboring position. We assume no observation noise when localizing to focus more on the effect of transition noise, except for the results in Section \ref{sec:results:rl_planners_with_different_noises}.

\textbf{Training Process.} We trained each planner in the \texttt{train} environment, allotting 8 hours for \BASE{} and 12 hours for \ALG{} to reach ~200 episodes. Each model was trained using a GTX 2080 Ti. The parameters of the \BASE{} and \ALG{} agents are defined in Table \ref{table:op_parameters}.

\begin{figure}
    \centering
    \includegraphics[width=\linewidth]{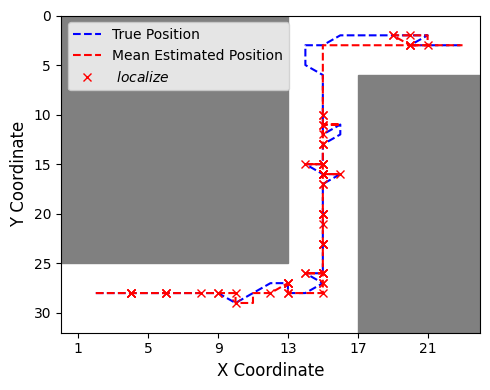} 
    \caption{A qualitative example of robot navigation: \textbf{blue}) actual path, \textbf{red}) estimated path and localization positions.} 

    \label{fig:Qualitative Example}
\end{figure}

The following subsections compare the performance of our baselines and \ALG{}. Our experiments ran until the robot reached the goal or failure region. The results were based on 100 runs for each algorithm in each environment (except \texttt{CC-POMCP}, which only has 75 runs in the \texttt{train} environment). In the $\texttt{chesapeake}$ and \STT{} environments, we randomized the path for each iteration, where the start and goal points were randomly selected from two open spaces that were at least 50 blocks away from each other.

\subsection{\ALG{} Qualitative Example }

Figure \ref{fig:Qualitative Example} presents a qualitative example demonstrating the robot's behavior in a noisy and uncertain environment, characterized by a (80\%, 10\%, 10\%)  transition noise, (68\%, 4\%) observation noise, and 40\% risk threshold. The results show that the robot consistently localizes in the Start Area and Middle Tunnel to mitigate failing early. Additionally, the agent increases its localization frequency within the Middle Tunnel, where the risk of failure is higher and noise can cause deviations from the intended path. Since the area after the middle tunnel is more spacious and thus safer, the agent does not localize as often. Finally, the agent localizes near the goal to ensure precise positioning for successful task completion. This adaptive behavior highlights the agent's strategy to navigate effectively under challenging conditions. 

\subsection{Comparing with Baseline}\label{sec:results:all_algorithms}

\begin{figure}
    \centering
    \includegraphics[width=0.75\linewidth]{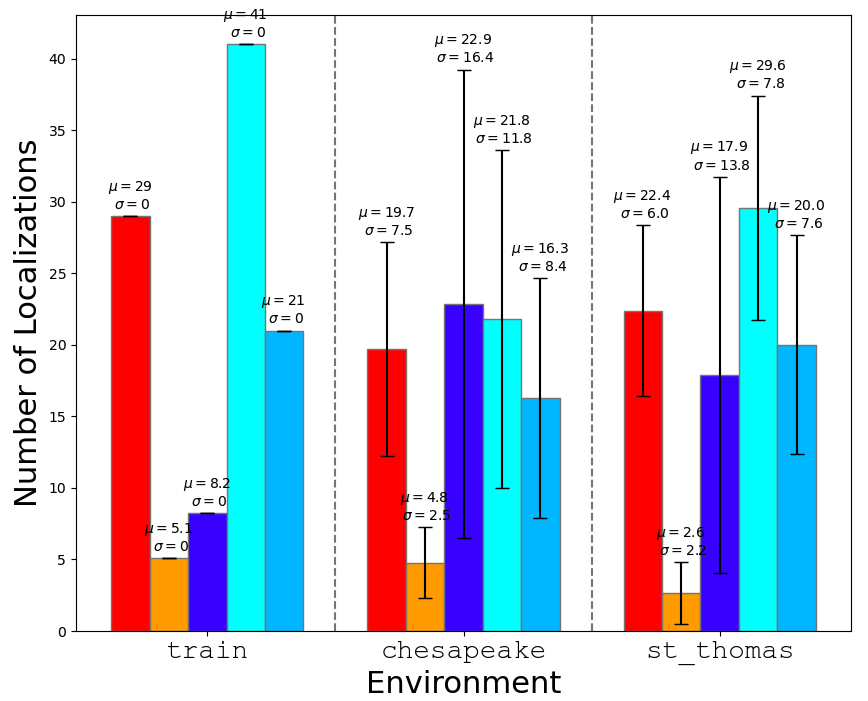}
    \label{fig:num_localize_bar}

    \includegraphics[width=0.75\linewidth]{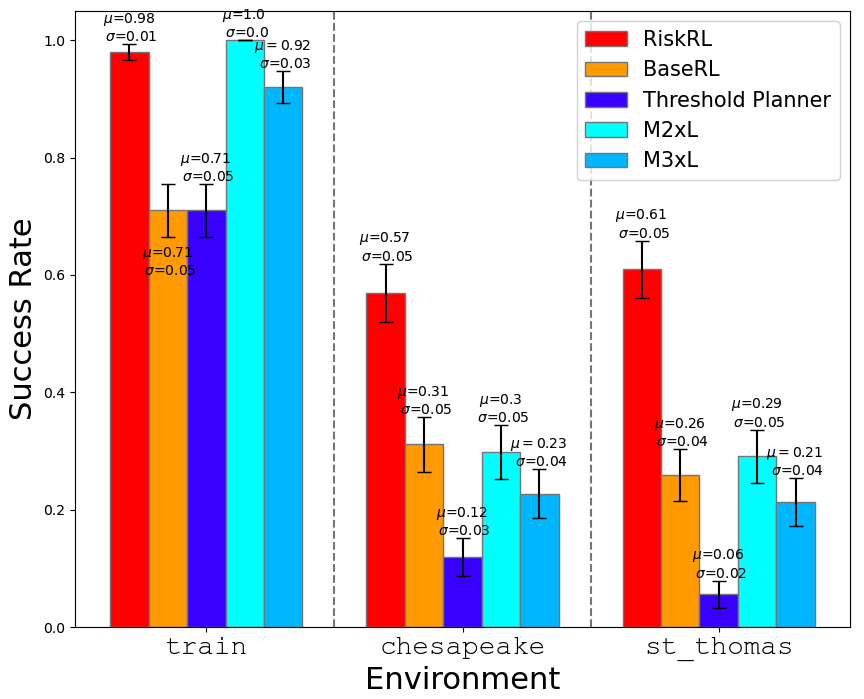}
    \label{fig:success_rate_bar}

    \caption{This figure depicts the number of localizations (top) and success rates (bottom) in each environment.} 
    \label{fig:num_and_success_bar}
\end{figure}

We compared all algorithms in terms of the number of localize actions and success rates (that is, reaching the goal). The \texttt{SP}, \texttt{TP}, and \texttt{CC-POMCP} results represent the average performance of each policy type. In our experiments, we set $c=0.4$ for general training, and the policy successfully achieves this threshold. Even though in the \texttt{train}, \ALG{} had similar success rates to \texttt{(M2xL)} and \texttt{(M3xL)} while still having a relatively high number of localizations, our experiments show that our \ALG{} planner generalized very well to the harder \texttt{chesapeake} and \texttt{st\_thomas} environments (Figure \ref{fig:num_and_success_bar}) as it was able to achieve a success rate nearly double of all of the other planners while keeping the number of localizations only marginally higher than \texttt{M3xL}. In the \texttt{chesapeake} environment, \ALG{} has a 57\% success rate while the next highest planner has a 31\% success rate while averaging 19.7 localizations. In the \STT{} environment, \ALG{} has a 61\% success rate while the next highest planner has a 29\% success rate while averaging 22.4 localizations. 

It is interesting to note that even though \BASE{} averaged very few localizations in the \texttt{chesapeake} and \STT{} environments (4.8 and 2.6, respectively), it was able to achieve a success rate on par with \texttt{M2xL}.

\begin{figure}
    \centering

    \includegraphics[width=0.75\linewidth]{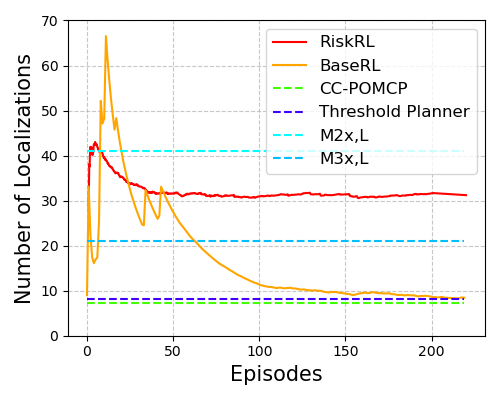}\\
    \includegraphics[width=0.75\linewidth]{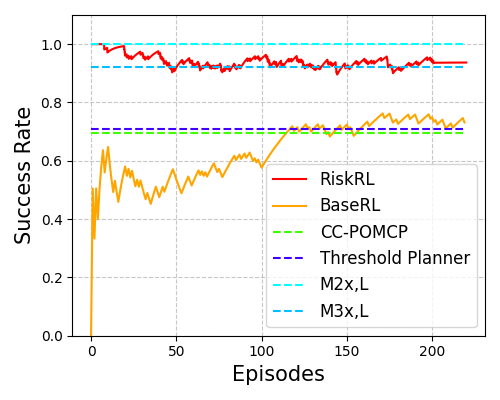}
    \caption{The training graphs for \texttt{train}, showing the number of localizations (top) and success rates (bottom) and $c$ set to 0.4.} 
    \label{fig:num_and_success_train}
\end{figure}

Compared to \texttt{CC-POMCP} (Figure \ref{fig:num_and_success_train}), \BASE{} performs similarly with a slightly higher success rate and number of localizations in the \texttt{train} environment. However, in terms of inference time, \texttt{CC-POMCP} takes around 20 seconds to choose one action, while \BASE{} and \ALG{} only take a maximum of 20 milliseconds.

\subsection{Analysis of RL Policies}
\begin{figure*}
    \centering
    \includegraphics[width=\linewidth]{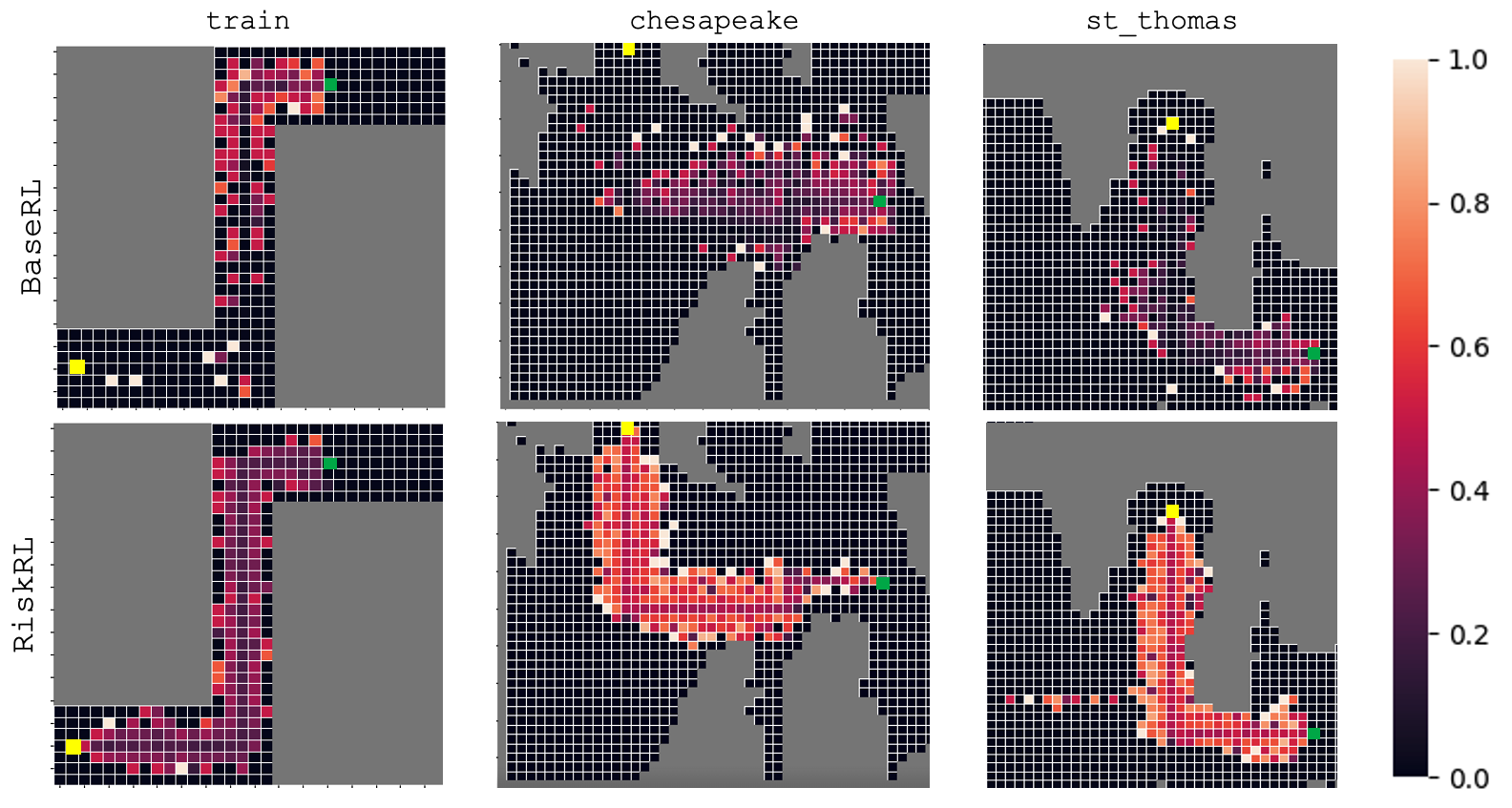}
    \caption{This figure shows the probability of localizing based on the agent's estimated location. The yellow block denotes the start position, and the green block denotes the goal position. Note for the $\texttt{chesapeake}$ and \STT{} environments, we only show a smaller segment of the environment. We ran each algorithm on each environment with a predefined path 250 times.} 

    \label{fig:HeatMaps}
\end{figure*}

Figure \ref{fig:HeatMaps} shows where each RL planner tends to localize based on the belief's mean. In general, \BASE{} localizes without considering the success constraint, leading the planner to localize more near obstacles or around turns where the percentage of particles collided is high. In contrast, \ALG{} distributes localizations more uniformly along the path with higher concentrations near obstacles, leading to a more robust policy with a higher success rate.

\subsection{Effect of Transition and Observation Noise}\label{sec:results:rl_planners_with_different_noises}

\begin{figure}
    \centering
    \includegraphics[width=0.75\linewidth]{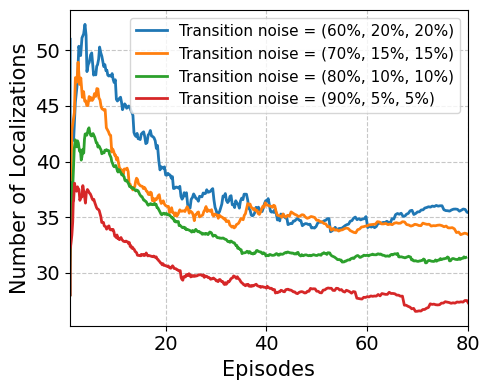}
    
    \includegraphics[width=0.75\linewidth]{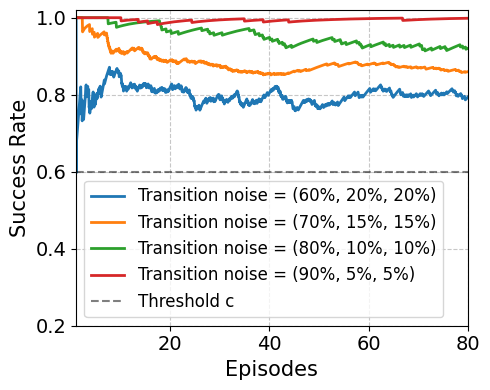}

    \caption{This figure shows the localization counts (top) and success rates (bottom) at different transition noises in \texttt{train}.} 
    \label{fig:Transition}
\end{figure}

\textbf{Varying Transition Noise.}
Figures \ref{fig:Transition} illustrate the number of localizations and success rates for \ALG{} as we vary the transition noise and use no observation noise. Here, the allowable probability of failure was set to $c=0.4$. Unsurprisingly, the localization count increases as the transition noise increases. More crucially, we observe that \ALG{} respects the constraint even at the highest transition noise, reaffirming the correctness of the algorithm. 

\begin{figure}
    \centering
    \includegraphics[width=0.75\linewidth]{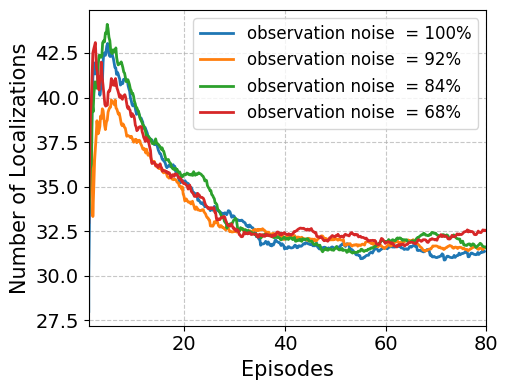}

    \includegraphics[width=0.75\linewidth]{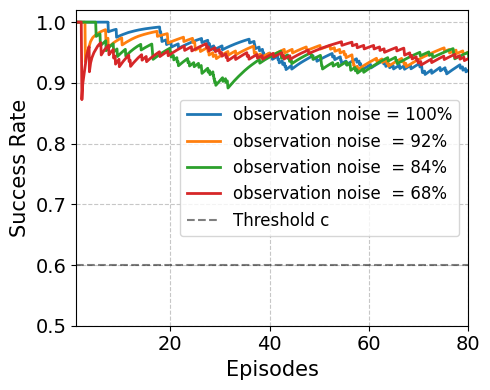}

    \caption{This figure shows the localization counts (top) and success rates (bottom) for different observation noises, where each percentage is the probability of observing the true position.} 
    \label{fig:Observation}
\end{figure}

\textbf{Varying Observation Noise.} We now study how the observation noise affects the performance of \ALG{} with the default transition noise.
Figures \ref{fig:Observation} demonstrate that observation noise does not impact the agent's performance. The results indicate that, despite increasing observation noise, the success rates remain relatively consistent across different levels. A similar trend is observed for the number of localizations, suggesting that our algorithm maintains robust performance in the presence of observation noise.

\begin{figure}
    \centering
    \includegraphics[width=0.75\linewidth]{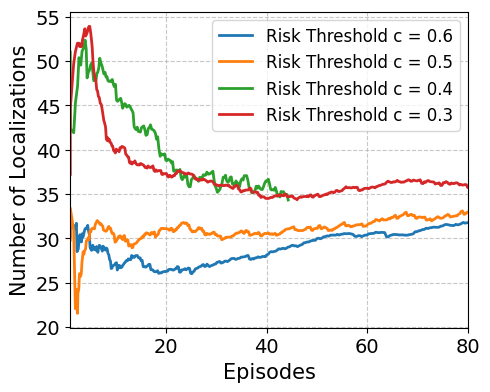}

    \includegraphics[width=0.75\linewidth]{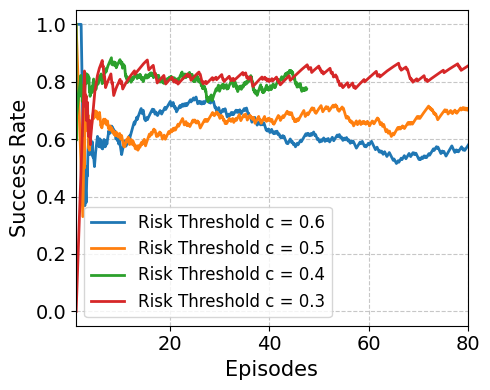}

    \caption{This figure shows the localization counts (top) and success rates (bottom) for varying risk thresholds $c$.} 
    \label{fig:Risk}
\end{figure}

\subsection{Effect of Risk Constraint}

Figures \ref{fig:Risk} demonstrate that the risk threshold $c$ influences the localization counts and success rates. As $c$ increases, the agent becomes more risk-averse, resulting in a higher frequency of localization and fewer failures. Conversely, with a lower risk threshold, the agent takes more risks, leading to increased uncertainty and a higher likelihood of failure. Consequently, the success rate diminishes as the risk constraint becomes more lenient. 

\section{Discussion}
The proposed \ALG{} framework balances localization frequency with risk constraints, demonstrating robust performance across different environments. While our approach generalizes well to unseen environments, its adaptability to significantly different domains (e.g., highly dynamic or adversarial settings) warrants further investigation. Additionally, although \ALG{} reduces inference time compared to CC-POMCP, training remains computationally expensive, requiring extensive interaction with the environment.

\section{Conclusion}\label{sec:conclusion}
We developed a novel active localization approach termed \ALG, which combines a high-level, chance-constrained planner with a particle filter (PF). Our approach aims to minimize localization actions while not exceeding failure probabilities. The chance-constrained planner determines when the robot moves or localizes and was implemented using constrained reinforcement learning (RL). The PF maintains the robot's belief, processing noisy motion commands and 2D pose observations from the environment. We also use the PF's belief to compute a 2D observation vector for the RL planner. The current approach succeeds our prior work \cite{Williams2024w2l}, which employed an algorithm that has a slower inference time and requires well-defined transition and observation noise models, making it unusable in real-time, real-world scenarios. Our results revealed three key findings. First, \ALG{} was able to achieve at least a 26\% higher success rate compared to the baselines when deployed in unseen test environments (\texttt{chesapeake} and \STT{}). This demonstrates the robot's ability to optimize when to localize, achieving higher rewards. Second, the robot dynamically adjusts its localization frequency, showcasing that it can adapt to different scenarios and environmental conditions. Finally, through bi-criteria optimization, \ALG{} effectively controls risk levels while maximizing performance, ensuring the robot operates safely. Notably, \ALG{} demonstrates zero-shot transfer capabilities, handling new environments without retraining, underscoring its potential for real-world deployment. 

Our future plans include further optimization to improve results (for example, by implementing \textit{Evolving Rewards} \cite{evolving}), exploring continuous state and action spaces, and evaluating in more realistic environments.

\addtolength{\textheight}{-11cm}   



\bibliographystyle{IEEEtran}

\end{document}